\documentclass[a4paper, 10pt, conference]{ieeeconf}      % Use this line for a4 paper

\IEEEoverridecommandlockouts                              % This command is only needed if 
\usepackage{graphics} % for pdf, bitmapped graphics files
\usepackage{epsfig} % for postscript graphics files
\usepackage{mathptmx} % assumes new font selection scheme installed
\usepackage{times} % assumes new font selection scheme installed
\usepackage{amsmath} % assumes amsmath package installed
\usepackage{amssymb}  % assumes amsmath package installed
\usepackage{balance}
\title{\LARGE \bf
Measuring a Soft Resistive Strain Sensor Array by Solving the Resistor Network Inverse Problem
}

\newcommand{\etal}{\textit{~et~al}. }

\usepackage{color}
\usepackage[dvipsnames]{xcolor}
\newcommand{\yz}{\textcolor{black}}

\author{Yuchen Zhao, Choo Kean Khaw, Yifan Wang% <-this % stops a space
\thanks{*
This work was supported by Singapore Agency for Science, Technology and Research (A*STAR) under its MTC IRG award M21K2c0118 and A*STAR AME YIRG award A2084c0162.
Y.W. would like to acknowledge Nanyang Technological University for its NAP award 020482.}% <-this % stops a space
\thanks{The authors are with School of Mechanical and Aerospace Engineering, Nanyang Technological University, 50 Nanyang Avenue, Singapore, 639798
        {\tt\small yifan.wang@ntu.edu.sg}}%
\thanks{
\textsuperscript\textcopyright 20XX IEEE. Personal use of this material is permitted. Permission from IEEE must be obtained for all other uses, in any current or future media, including reprinting/republishing this material for advertising or promotional purposes, creating new collective works, for resale or redistribution to servers or lists, or reuse of any copyrighted component of this work in other works.}
}

\begin{document}

\maketitle
\thispagestyle{empty}
\pagestyle{empty}

%%%%%%%%%%%%%%%%%%%%%%%%%%%%%%%%%%%%%%%%%%%%%%%%%%%%%%%%%%%%%%%%%%%%%%%%%%%%%%%%
\begin{abstract}

Soft robotics is applicable to a variety of domains due to the adaptability offered by the soft and compliant materials.
To develop future intelligent soft robots, soft sensors that can capture deformation with nearly infinite degree-of-freedom are necessary.
Soft sensor networks can address this problem, however, measuring all sensor values throughout the body requires excessive wiring and complex fabrication that may hinder robot performance.
We circumvent these challenges by developing a non-invasive measurement technique, which is based on an algorithm that solves the inverse problem of resistor network, and implement this algorithm on a soft resistive, strain sensor network.
Our algorithm works by iteratively computing the resistor values based on the applied boundary voltage and current responses, and we analyze the reconstruction error of the algorithm as a function of network size and measurement error.
We further develop electronics setup to implement our algorithm on a stretchable resistive strain sensor network made of soft conductive silicone, and show the response of the measured network to different deformation modes.
Our work opens a new path to address the challenge of measuring many sensor values in soft sensors, and could be applied to soft robotic sensor systems.

\end{abstract}

%%%%%%%%%%%%%%%%%%%%%%%%%%%%%%%%%%%%%%%%%%%%%%%%%%%%%%%%%%%%%%%%%%%%%%%%%%%%%%%%
\section{INTRODUCTION}
Robots made of soft and compliant materials are shown to be capable of various tasks~\cite{rus_2015_design,shintake_2018_soft,goh_2022_3d}, such as adaptive locomotion~\cite{shepherd_2011_multigait}, universal gripping~\cite{brown_2010_universal}, and are resilient to damage~\cite{tolley_2014_resilient}.
Sensing ability can also make robots more intelligent, and seamless integration of novel soft actuators, sensors and control systems is critical for developing next-generation intelligent robots for everyday yet demanding tasks such as shape sensing, object detection and classification~\cite{meerbeek_2018_soft,hughes_2018_tactile,zhang_2021_fruit}.

Equipping soft robots with sensing abilities provides new challenges.
Not only the sensors need to be compatible with the high deformability of the robots during operation, but also distributed in many part of the continuous body to capture the potentially infinite degree-of-freedom deformation~\cite{hughes_2018_tactile}.
Developments in flexible electronics can address this issue~\cite{souri_2020_wearable,shih_2020_electronic,castano_2014_smart}.
Various soft sensors have been developed, such as 
conductive ink containing carbon black is printed onto flexible textile to make sensors ~\cite{buckner_2020_roboticizing}
Sliver particle ink is printed onto plastic sheet to form tough pad~\cite{nittala_2018_multi-touch}.
Serpentine structures are fabricated out of rigid materials, resulting very flexible arrays for embedding different sensors~\cite{hua_2018_skin-inspired}.
Resistive strain sensors are integrated into soft pneumatic bending actuator to detect bending and pressure states~\cite{pinto_2017_cnt-based,koivikko_2018_screen-printed}.
Sensors that are only sensitive to pressure but not bending are fabricated~\cite{lee_2016_transparent}.
3D printing of conductive elastomer is used to fabricate fin ray grippers for sensing and gripping~\cite{georgopoulou_2022_pellet-based}.
The infinite possibility of deformation due to the continuous body poses another challenge for developing sensor systems for robotic intelligence.
Adding more soft sensor is a natural choice, but measuring signals from sensors throughout the robot body is a challenge.
In addition, the abundant sensory information may require novel acquisition and processing methods to uncover the mechanical deformation of the body.

In a step towards soft sensors for complex deformation patterns, we investigate a soft, resistive strain sensor network and develop algorithms and measurement techniques to retrieve individual sensor signals only via peripheral measurements.
Soft sensors based on smart materials can easily adopt different array or network geometry during fabrication.
One common approach is to organize sensors into an array with orthogonal transmitter and receiver lines scanning through each sensor sequentially~\cite{pourjafarian_2019_multi-touch}.
Soft capacitive sensor array of this kind is tested on universal jamming gripper and wearable device~\cite{audenaert_2015_granual,loh_2021_3d}.
Besides directly reading sensor values, proper modelling of the interaction between sensors and its environment can provide valuable information for robot intelligence.
Hughes\etal uses conductive elastomeric wires stitched onto the surface of a universal jamming gripper to retrieve object location, shape and force information~\cite{hughes_2018_tactile}.
By assuming the object being a sphere with unknown center coordinates and radius, the conformation of the object to the gripper leads to gripper surface deformation and wire straining.
The relation between the different amount of wire straining and the unknowns can be established and solved numerically.
In addition, machine learning (ML) techniques are used to connect sensor information to complex deformation of the soft body.
For example, Meerbeek\etal develops an internally illuminated elastomer foam via embedded optical fibers that can detect its own deformation through ML techniques~\cite{meerbeek_2018_soft}.
The optical fibers transmit light into the foam, and light diffusion in the foam is related to the mechanical deformation.
The diffused light is then transmitted back via the same fibers and captured by a camera.
Data are collected for different deformation and light intensity, and ML models are trained to capture the relation between the two.
Also, Hughes\etal uses ML to relate sensor array signals to object features on a gripper where a granular jamming compartment is placed between the object and the sensor to allow changeable sensor morphology~\cite{hughes_2021_online}.

Ideally, the embedded sensor network should have a minimal influence on the deformation of the body and be able to obtain the accurate deformation.
Many current sensor networks requires additional effort for wiring needs, since each sensor signal needs to be read independently.
This approach could be limited by the material choice, sensor shape and sensor density.
Directly fabricating these sensors as connected can lift some of the limitation mentioned above.
However, the interconnection between these sensors bring additional challenges for signal processing.
For example, Medina\etal studies a flexible resistive sensor network with distributed sensors that response differently to mechanical deformation~\cite{medina_2017_resistor-based}.
By applying pairs of voltages at peripheral nodes and measuring the response of the rest, the resistor values can be estimated for all sensors in the network via a computer tomography-inspired optimization algorithm.
In fact, resistor network inverse problem has been studied and various methods are developed~\cite{curtis_2000_inverse,oberlin_2000_discrete,borcea_2008_electrical}, but as far as we know, no experimental prototype is demonstrated.
% Our work
Based on the work of Curtis and Morrow~\cite{curtis_2000_inverse}, we develop an iterative algorithm that is applicable to solve the inverse problem of resistor network.
We test our algorithm on a soft resistive strain sensor network, and demonstrate that the spatial distribution of relative resistance change of individual resistor segment is correlated to the different mechanical deformations applied to the sensor network.
We further study the capacity of our inverse problem algorithm via numerically simulate resistor network of different sizes and measurement noise.
Since only peripheral electric currents measurement are needed, the network design can be scaled easily without wiring concerns, and various existing fabrication techniques can be directly applied to fabricate sensor networks.
\yz{Existing reconstruction algorithms and systems based on electrical impedance tomography (EIT) and ML techniques~\cite{park2022biomimetic,shih2020electronic} are fast-developing.
However, typical EIT algorithms are slow for real-time applications~\cite{park2022biomimetic}. ML techniques require large datasets for training, and the relationship between measurement value and mechanical response is usually a black box to researchers. 
Our algorithm allows all resistor values in the network to be measured simultaneously in a short time (on the order of milliseconds) and the method is interpretable, making our method transferable to all kinds of resistive-type sensor networks.}

Our paper is organized as follows. Sec.~\ref{sec:fab} describes the materials and fabrication steps for the soft resistive strain sensor network.
In Sec.~\ref{sec:math} we write down the mathematical formulation for the resistor network and the algorithm to solve the inverse problem.
In Sec.~\ref{sec:exp} we describe the measurement electronics circuit and show reconstruction results of the soft resistor network sensor under different stretching condition.
Lastly, we report numerical simulations of resistor network and the accuracy of the inverse problem algorithm in Sec.~\ref{sec:simu} 

\section{SENSOR MATERIALS AND FABRICATION}\label{sec:fab}
\begin{figure}[thpb]
\centering
\framebox{\parbox{3.2in}{\includegraphics[width=3.2in]{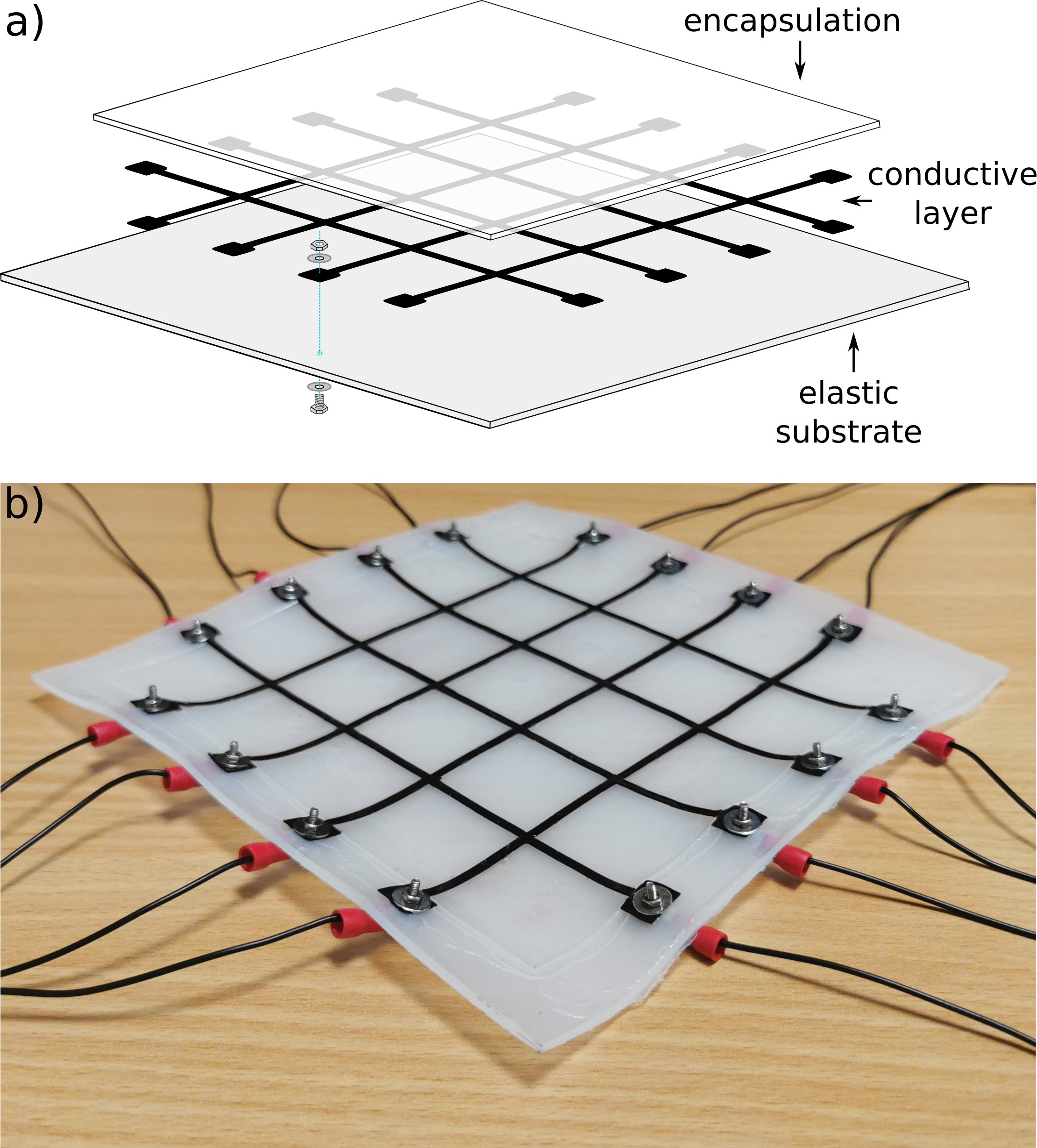}}}
\caption{Illustration of the stretchable resistor network sensor. a) a schematics of the sensor, including a silicone elastomer substrate, conductive silicone rubber for sensing, and silicone glue for encapsulation.
Washer, bolts and nuts are used to secure electric contact with the electrode pad from the edge of the resistor network.
b) a picture of the sensor.}
\label{fig:fab}
\end{figure}

Our stretchable resistor network sensor consists of three layers of silicone sheet, which are commercially-available, non-toxic and more eco-friendly than plastics.
Fig.~\ref{fig:fab} illustrate its composition.
A square sheet of silicone elastomer (MALAYSIACLAYART 45A) of size $19\,$cm and thickness $2\,$mm is used as the stretchable substrate to hold the sensor.
A conductive rubber sheet (3bluesky-alina, Aliexpress.com) of thickness $0.5\,$mm is cut into a square grid pattern with four strips on each sides(Fig.~\ref{fig:fab}a).
This results in $40$ line segments, each of which is a resistive strain sensor those resistance increases when elongated.
The length and width of each line segment is $3\,$cm and $0.2\,$mm, respectively, and the resistance is between $400$ to $700\,\Omega$, likely due to the observed anisotropic conductivity in the sheet.
The line segments together form the stretchable resistor network.
To facilitate a good and durable contact at the network's edges, we left a square pad of $8\,$mm at the boundary edge.
The cut network is bonded to the elastic substrate by a thin layer of silicone glue (Kafuter K5905).
Then, we encapsulate the sensor with another transparent silicone glue (Kafuter K704) of thickness about $0.7\,$mm, except the square pads.
The exposed square pads serve as the electrodes of the resistor network, and we use metal washers, bolts and nuts to secure the electrical contact with fork terminal wires (Fig.~\ref{fig:fab}b).

\section{INVERSE PROBLEM OF RESISTOR NETWORK}\label{sec:math}
\begin{figure}[thpb]
\centering
\framebox{\parbox{3.2in}{\includegraphics[width=3.2in]{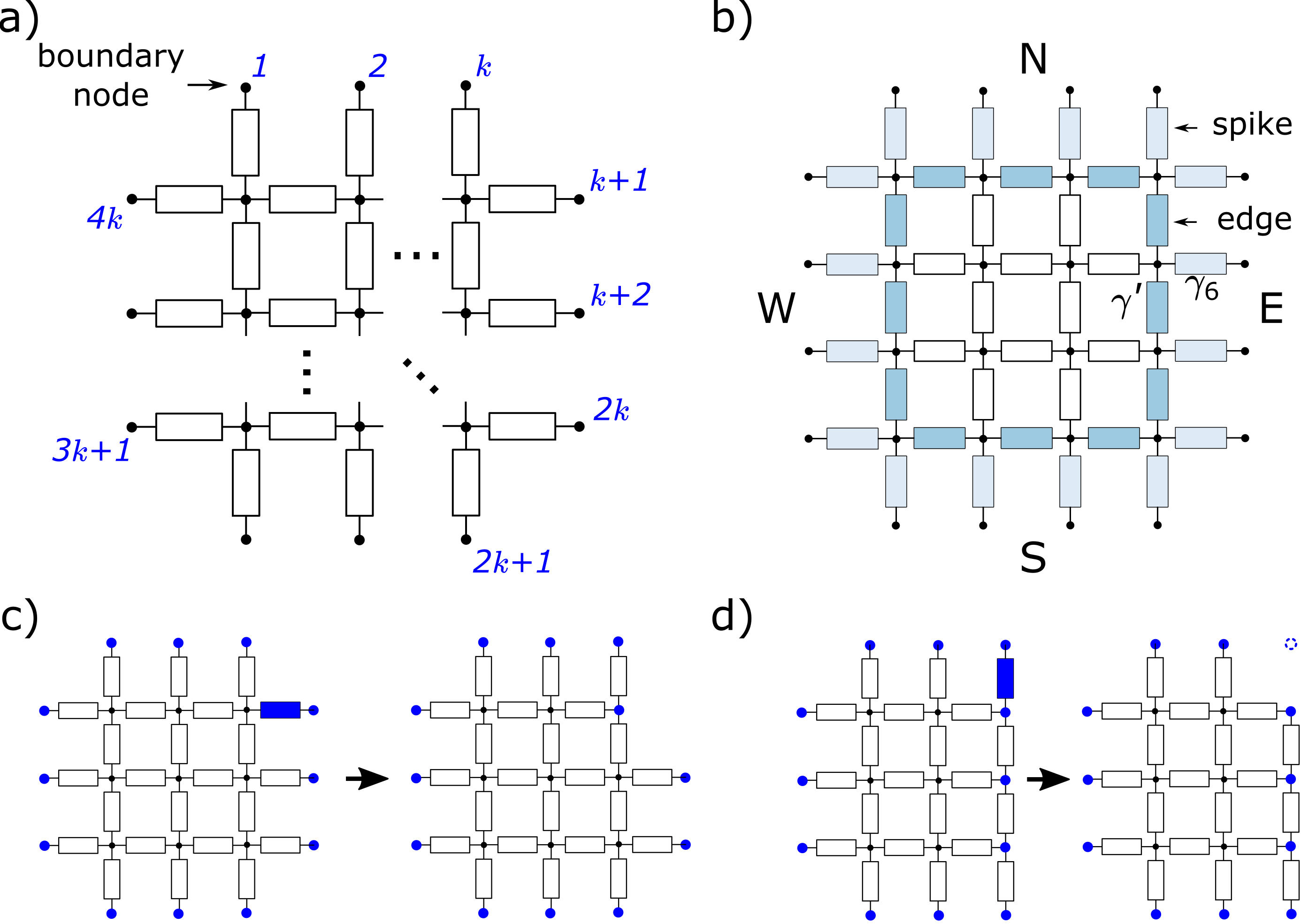}}}
\caption{The resistor network model.
(a) A $k$-by-$k$ square lattice resistor network.
Boundary nodes are indexed from $1$ to $n=4k$ clockwise, starting from the top-left corner.
(b) A 4-by-4 resistor network those boundary spike and edge resistors are highlighted in different colors.
N, E, S, W stand for the north, east, south and west face of the network.
(c) Illustration of the boundary spike resistor removal process.
All boundary nodes, whether before or after the transformation, are highlighted in big blue dots.
(d) Illustration of the boundary edge resistor removal process.
After removal, the original boundary node in the top-right corner becomes isolated to the rest of the network.
The corresponding columns and rows in the $\Lambda$ becomes zero.
}
\label{fig:resistor_network}
\end{figure}

We consider a square resistor network model as illustrated in Fig.~\ref{fig:resistor_network}.
The resistors can have different values. 
This model is used as an approximation to the resistor network sensor we fabricated.
The length of the network $k$ is given by the number of nodes on its edge.
A resistor network of length $k$ has a total of $n=4k$ boundary nodes.
For simplicity, we let the node in the top-left corner to be the first node, and enumerate the rest nodes clockwise, as shown in Fig.~\ref{fig:resistor_network}a.
Following the notation in ref.~\cite{curtis_2000_inverse}, when we apply a set of voltages $\mathbf{u}=(u_1,u_2,\cdots,u_n)$ at all boundary nodes, electric current will flow through the resistor network according Kirchhoff circuit law.
Let $\pmb{\phi}=(\phi_1,\phi_2,\cdots,\phi_n)$ be the collection of currents that flows through the boundary nodes, the relation between $\mathbf{u}$ and $\pmb{\phi}$ is determined by the $n$-by-$n$ response matrix $\Lambda$ as:
\begin{eqnarray}
\pmb{\phi} = \Lambda \mathbf{u}
\end{eqnarray}
\noindent assuming each resistor follows the Ohm's law.
The response matrix is symmetric, and is used to reconstruct all resistor values in the resistor network~\cite{curtis_2000_inverse}, and measurement of this response matrix is the starting point of our reconstruction algorithm.
Entries in $\Lambda$ has the unit of conductance.
The $i$'th column of $\Lambda$ can be interpreted as the values of the currents flow through boundary nodes when a unit voltage is applied to the $i$'th boundary node, and zero voltage is applied to all the rest boundary nodes.
This physical interpretation is the basis of our measurement method.
By sequentially applying this singular voltage boundary condition to all boundary nodes and measure the current values through all nodes, we can obtain the response matrix $\Lambda$.

Our reconstruction algorithm uses a combination of the method of special functions and network adjunction, both described in the book~\cite{curtis_2000_inverse} to resolve the values of internal resistors in the network.
The method of special functions constructs special boundary conditions and `propagate' these boundary values via harmonic continuation, so that the boundary resistor values can be calculated via simple formula based on elements of $\Lambda$.
The network adjunction method can reduce the resistor network $\Lambda$ by removing the known resistors on the boundary.
These boundary resistors are highlighted in Fig.~\ref{fig:resistor_network}b, and are further classified into two types.
The boundary spike resistors are those connecting boundary nodes to the interior of the networks, and the boundary edge resistors are those joining two boundary spike resistors.
We uses the two methods in an alternating fashion, described below:
\begin{itemize}
\item \textit{Step 1}: Given an n-by-n response matrix $\Lambda_{n \times n}$, compute all boundary spike resistors and boundary edge resistors.
These resistors are highlighted in Fig.~\ref{fig:resistor_network}b.
\item \textit{Step 2}: Using the calculated boundary spike and edge resistor values, reduce $\Lambda_{n \times n}$ to $\Lambda_{n-8 \times n-8}$.
The latter corresponds to the resistor network of the original one when all boundary spike and edge resistors are remove (the minus $8$ comes from the four faces of the network with two end resistors to be removed for each face).
\item \textit{Step 3}: Repeat \textit{Step 1} for $\Lambda_{n-8 \times n-8}$ until all resistor values are computed.
\end{itemize}

To calculate the boundary spike and boundary edge resistors, we following the method of special functions described in ref.~\cite{curtis_2000_inverse}.
The results are stated below.
We first write $\Lambda$ in block form as:
\begin{eqnarray}
\Lambda = 
\begin{pmatrix}
\Lambda_{NN} & \Lambda_{NE} & \Lambda_{NS} & \Lambda_{NW}\\
\Lambda_{EN} & \Lambda_{EE} & \Lambda_{ES} & \Lambda_{EW}\\
\Lambda_{SN} & \Lambda_{SE} & \Lambda_{SS} & \Lambda_{SW}\\
\Lambda_{WN} & \Lambda_{WE} & \Lambda_{WS} & \Lambda_{WW}
\end{pmatrix}
\end{eqnarray}
\noindent where `N', `S', `E', `W', stand for north, south, east and west, indicating the differently oriented faces of the resistor network (Fig.~\ref{fig:resistor_network}b).
The following matrices are calculated:
\begin{equation}
\begin{aligned}
\tilde{\Lambda}_N &= \Lambda_{NN} - \Lambda_{NE} \cdot \Lambda_{WE}^{-1} \cdot \Lambda_{WN}\\
\tilde{\Lambda}_E &= \Lambda_{EE} - \Lambda_{EN} \cdot \Lambda_{SN}^{-1} \cdot \Lambda_{SE}\\
\tilde{\Lambda}_S &= \Lambda_{SS} - \Lambda_{SW} \cdot \Lambda_{EW}^{-1} \cdot \Lambda_{ES}\\
\tilde{\Lambda}_W &= \Lambda_{WW} - \Lambda_{WS} \cdot \Lambda_{NS}^{-1} \cdot \Lambda_{NW}
\end{aligned}
\end{equation}
\\
The conductance $\gamma_i$ of the boundary spike resistor connecting to boundary node $i$ equals to the diagonal element of the corresponding $\tilde{\Lambda}_{\,*}$,
and the conductance $\gamma^{\;\prime}$ of the boundary edge resistor connected to $i$ and $i+1$ boundary nodes equals to the product of $\gamma_i$ and $\gamma_{i+1}$ divided by the off-diagonal element $(\tilde{\Lambda}_{\,*})_{i,i+1}$.
For example, $\gamma_6$ in the 4-by-4 network shown in Fig.~\ref{fig:resistor_network} equals to $(\tilde{\Lambda}_E)_{22}$, and $\gamma^{\;\prime}$ that is between $\gamma_6$ and $\gamma_7$ equals to $\gamma_6 \gamma_7 / (\tilde{\Lambda}_E)_{23} = (\tilde{\Lambda}_E)_{22} (\tilde{\Lambda}_E)_{33} / (\tilde{\Lambda}_E)_{23}$.

The method of special functions does not provide a simple close-form expression to calculate all resistors that are inside the network.
We overcome this problem by applying resistor adjunction and removal~\cite{curtis_2000_inverse}, so that the inverse problem of a network of length $k$ can be reduced to a network of length $k-2$.
The adjunction method applies a transformation on $\Lambda_{n \times n}$ so that the resulting matrix corresponds to a new network with one boundary resistor (either a boundary edge or a boundary spike) removed from the original one.
Since all boundary spike and edge resistor values are known, we remove these resistor one-by-one until the resulting network is similar to the original one except smaller in length (Fig.~\ref{fig:resistor_network}b colorless network).
This procedure translates into a series of transformation on the $\Lambda_{n \times n}$.
Some columns and rows will become zero since some original nodes are not connected to anything in the resulting network, so that $\Lambda_{n \times n}$ can be reduced to a smaller matrix.
In particular, two kinds of transformation methods are used, namely removing a boundary edge resistor (Fig.~\ref{fig:resistor_network}c) and a boundary spike resistor (Fig.~\ref{fig:resistor_network}d).
Without loss of generality, assuming the boundary spike resistor to be removed is at the $i=1$, and it has a conductance of $\gamma$.
Then the following transformation applies:
\begin{eqnarray}\label{eqn:spike_rm}
\Lambda = 
\begin{pmatrix}
\Lambda_{11} & \pmb{\lambda}^{T}\\
\pmb{\lambda} & \mathbf{C}
\end{pmatrix}
\quad
\rightarrow
\quad
S_{\gamma}(\Lambda)=
\begin{pmatrix}
-\gamma-\frac{\gamma^2}{\delta} & -\frac{\gamma}{\delta}\pmb{\lambda}^{T}\\
-\frac{\gamma}{\delta}\pmb{\lambda} & \mathbf{C} - \frac{\pmb{\lambda}^{T}\pmb{\lambda}}{\delta}
\end{pmatrix}
\end{eqnarray}
\noindent where $\delta = \Lambda_{11}-\gamma$ and $\pmb{\lambda}$ is a column vector.
The transformed matrix $S_{\gamma}(\Lambda)$ is the response matrix of the modified network when the resistor $\gamma$ is removed, and the interior node $\gamma$ connected to becomes the new boundary node, as illustrated in Fig.~\ref{fig:resistor_network}(c).
For boundary edge resistor removal, assuming that the resistor of conductance $\gamma^{\;\prime}$ connects boundary nodes $i$ and $j$.
The resulting matrix $T_{\gamma}(\Lambda)$ is :
\begin{equation}\label{eqn:edge_rm}
\begin{aligned}
T_{\gamma}(\Lambda)_{ii} &= \Lambda_{ii} - \gamma^{\;\prime}\\
T_{\gamma}(\Lambda)_{jj} &= \Lambda_{jj} - \gamma^{\;\prime}\\
T_{\gamma}(\Lambda)_{ij} &= \Lambda_{ij} + \gamma^{\;\prime}\\
T_{\gamma}(\Lambda)_{ji} &= \Lambda_{ji} + \gamma^{\;\prime}\\
T_{\gamma}(\Lambda)_{*,*} &= \Lambda_{*,*} \quad\text{otherwise}
\end{aligned}
\end{equation}
\noindent and the process is illustrated in Fig.~\ref{fig:resistor_network}(d).
Equations (\ref{eqn:spike_rm}) and (\ref{eqn:edge_rm}) do not change the size of $\Lambda$.
However, by removing resistors from the network, some boundary nodes will become isolated (connecting to no resistor), so the corresponding entries in $\Lambda$ will be zero.
Removing these zero rows and columns will reduce the $\Lambda$ to the right size, which concludes the step 2 of the algorithm.

\section{EXPERIMENTAL RESULTS}\label{sec:exp}
\subsection{Setup}\label{sec:exp-setup}
\begin{figure}[thpb]
\centering
\framebox{\parbox{3.2in}{\includegraphics[width=3.2in]{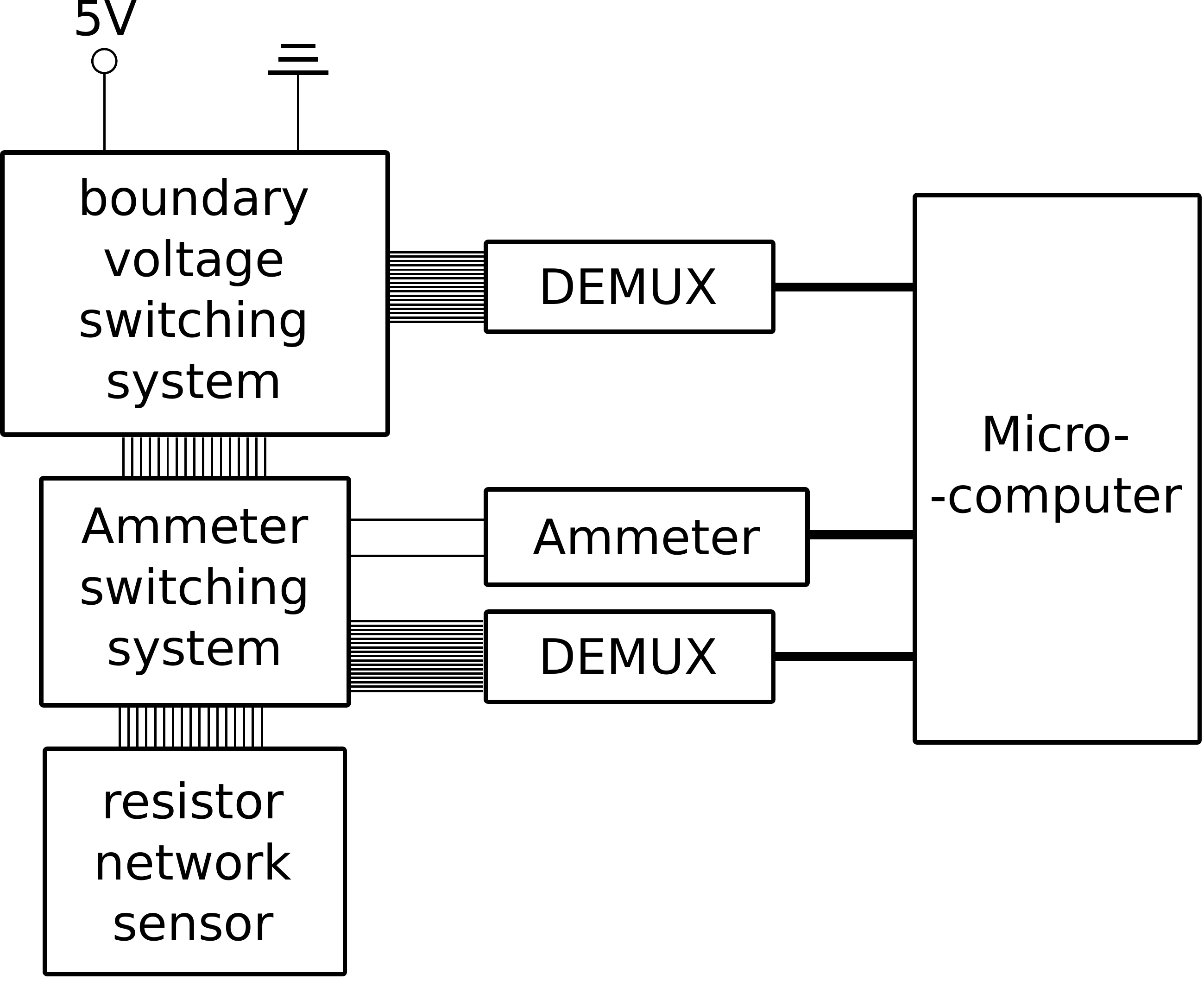}}}
\caption{A schematics of the electric circuit that measure the response matrix $\Lambda$ of the resistor network sensor.}
\label{fig:exp_setup}
\end{figure}

We design an electric circuit to measure the response matrix $\Lambda$ of the soft resistive strain sensor network and compute the resistor values under the square resistor network model.
The block diagram of the circuit is illustrated in Fig.~\ref{fig:exp_setup}.
There are three main components: a boundary voltage switching system to apply the singular-voltage boundary condition to the network, an ammeter switching system to insert one ammeter into individual boundary wire to measure the current flow, and the ammeter which is a voltage divider-based circuit measurement device.
For each electric line comes out of the boundary nodes, we apply a voltage of $0\,$V or $5\,$V using a common electro-magnetic relay (JQC-3FF-S-Z) as a single-pole double-throw switch.
The switching of the many relays is achieved by cascading demultiplexers (snx4hc138).
Because we only need one boundary node to connect to $5\,$V at a time, the cascaded demultiplexers module (DEMUX) conveniently convert a few bits input from the microcomputer to an array of digital signals that are forward to the relays.
In between the voltage switching system and the sensor is the ammeter switching module.
We use two relays to act as a double-pole double-throw switch that connect a boundary wire to either the ammeter or just shorted.
The switching of these relays is controlled by another DEMUX module.
The ammeter consists of a constant shunt resistor and a common voltage amplifier breakout board (HX711, 24-bit ADC) to measure the voltage across the shunt resistor, those resistance is chosen so that a maximum voltage around $20\,$mV across the shunt resistor can be achieved during all measurements.
\yz{The SNR, defined as the ratio of mean to standard deviation of repeated measurement $\mu/\sigma$, is about $230$ and $650$ for metal film resistors and the soft silicone rubber resistor network, respectively}.
The control of switching modules, ammeter readings processing, and reconstruction computation are carried out by a microcomputer (Raspberry Pi 4B).
  
In a typical $\Lambda$ measurement, we apply $5\,$V at one boundary node and the rest $0\,$V.
We then insert the ammeter next to each boundary node to measure the current flow.
The resulting current values correspond to a column in the $\Lambda$ matrix, and by switching the $5\,$V to other nodes, we can measure all columns in $\Lambda$.
Two cautions are taken during practice: 
1) the current passing through the $5\,$V boundary node is not directly measured, but given as a sum of all the other currents passing through the boundary nodes because the current is conversed.
Since this current is usually larger than the rest, by avoiding its measurement we can choose a larger gain in the ammeter for higher measurement resolution.
2) Theoretically speaking, $\Lambda$ is a symmetric matrix, but noise in measurements make it asymmetric.
We symmetrize $\Lambda$ by taking an average of its transpose before the reconstruction step.
The reconstruction of resistance values from $\Lambda$ is performed on the microcomputer according to the algorithm in Sec.~\ref{sec:math}.
%We further benchmark our circuit setup with common metal film resistors of known values, and successful reconstructions are tested with resistances ranging from $220\,\Omega$ to $22\,\text{k}\Omega$ and network length up to $5$ (results not shown).

\yz{
We further evaluate the reconstruction accuracy of our circuit setup on resistor networks consist of identical common metal film resistors.
The average value of the resistors are $22.08\pm0.04\,\text{k}\Omega$.
We find that the accuracy of the reconstruction depends on the value of the shunt resistor used in the ammeter.
We adjust the the shunt resistor such that the maximum voltage across the resistor is around $20\,$mV for different network sizes, and the relative RMSE of reconstructed resistance for a $2\times 2$, $3\times 3$, $4\times 4$, and $5\times 5$ network is $2.9\%$, $2.6\%$, $2.3\%$, and $12.7\%$, respectively. 
}

\subsection{Stretching test and resistor network reconstruction}\label{sec:exp-res}

\begin{figure}[thpb]
\centering
\framebox{\parbox{3.2in}{\includegraphics[width=3.2in]{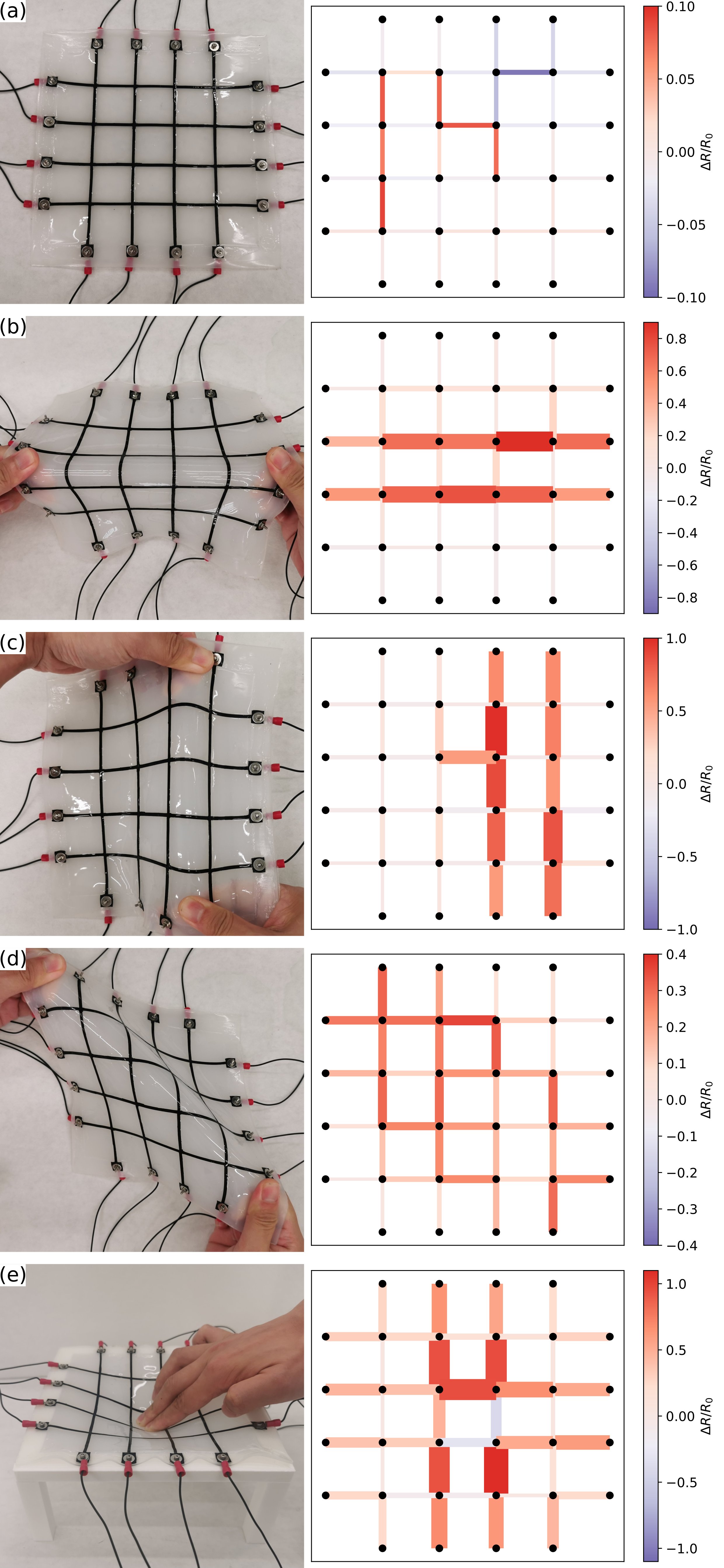}}}
\caption{
The soft resistive strain sensor network under different stretching modes and the corresponding distribution of relative resistance change $\Delta R/R_0$.
(a) unstretched state; (b) horizontal stretch; (c) vertical stretch; (d) diagonal stretch; \yz{(e) radial stretch}.
The left figure demonstrates the stretching mode.
The right figure shows the distribution of relative resistance change $\Delta R/R_0$.
Line colors and thickness are proportional to $\Delta R/R_0$.
Red/blue corresponds to positive/negative $\Delta R/R_0$.
Note that different stretching modes have different $\Delta R/R_0$ scale.
}
\label{fig:exp_result}
\end{figure}

We subject the resistor network sensor to different stretching modes and quantify the relative changes of its resistors, as shown in Fig.~\ref{fig:exp_result}.
We first measure $\Lambda$ twice in the unstretched rest state, and calculate the change in relative resistance $\Delta R/R_0$ for each resistor.
The \yz{best-achievable} fluctuation between different $\Lambda$ measurements is less than $10\%$ in $\Delta R/R_0$.
For the stretching tests, we first measure a $\Lambda_0$ in the unstretched state, then measure another $\Lambda$ when the sensor is under constant stretch.
\yz{The reconstructions were done offline}.
The resulting $\Delta R/R_0$ distribution is visualized as a color map shown in Fig.~\ref{fig:exp_result}(b), (c), (d), \yz{and (e)} under horizontal, vertical, diagonal, \yz{and radial} stretch.
Clear pattern in the $\Delta R/R_0$ distribution can be seen for different stretches, which is correlated to the physical stretch performed on the sensor.
For horizontal, vertical, and \yz{radial} stretch, the $\Delta R/R_0$ \yz{goes up to nearly $100\%$.}
For the diagonal stretch, $\Delta R/R_0$ is about $40\%$.
This result clearly indicates that our sensor is capable of detecting different stretching modes via the distribution of $\Delta R/R_0$ in the computed resistor network, and the measurement requires no invasive instrument or wiring, only currents at the boundary are measured.

\section{COMPUTER EXPERIMENTS OF RECONSTRUCTION ALGORITHM}\label{sec:simu}
\begin{figure}[thpb]
\centering
\framebox{\parbox{3.2in}{\includegraphics[width=3.2in]{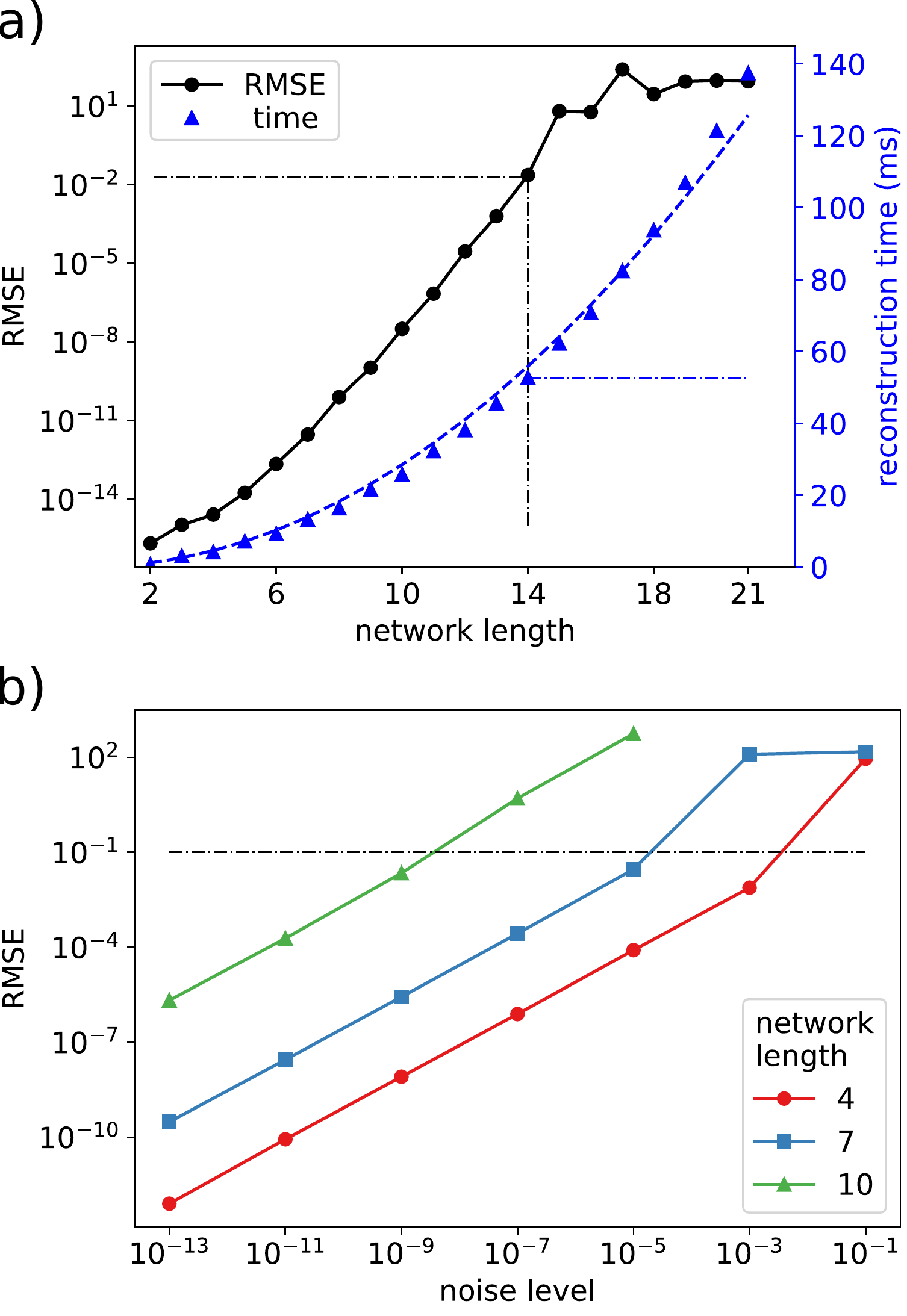}}}
\caption{Numerical simulation of the resistor network. (a) simulation error, defined as the RMSE (black circles and line) between the reconstructed and ground truth values of the resistance, and reconstruction time (blue triangles) are plotted as a function of network length.
The dashed blue line is a quadratic fit to the data.
Thin dash-dotted lines indicate the performance of network length equals $14$.
(b) reconstruction error is plotted as a function of noise level in the $\Lambda$ measurement for different network sizes, 4 (red circles), 7 (blue squares) and 10 (green triangles).
Thin dash-dotted line indicates an error equals $0.1$.
Each data point in (a) and (b) is an average over $100$ independent simulated networks.}
\label{fig:simu_result}
\end{figure}

To further explore the capacity of our reconstruction algorithm, we conduct computer experiments of the square resistor network and compare reconstructed resistances with ground truth values.
We start with a square resistor network shown in Fig.~\ref{fig:resistor_network} and all of its resistor values are known.
Its response matrix $\Lambda$ can be computed via the Kirchhoff matrix~\cite{curtis_2000_inverse}, and we use the same reconstruction algorithm in Sec.~\ref{sec:math} to compute all resistor values.
All computer simulations are carried out on a $1.5\,$GHz CPU on the microcontroller (Broadcom BCM2711, Quad core Cortex-A72).
The initial resistances are taken randomly between $1$ and $2$ (arbitrary unit), and we quantify the error between the ground truth and reconstructed values by computing the RMSE.
The RMSE is plotted as a function of network length in Fig.~\ref{fig:simu_result}(a), and for each network length, we average RMSE over $100$ independently generated resistor networks.
We see an exponentially growth of the reconstruction error.
When the network is small, the reconstruction error is comparable to the numerical precision we used in the calculation ($10^{-16}$).
We think that the exponential growth of the error comes of the iterative procedure of the algorithm, as the error presented in $\Lambda_{n\times n}$, even if very small, will propagate to $\Lambda_{n-8\times n-8}$ during the reduction step.
The error reaches about $0.01$ when the network length is $14$, indicating an upper limit on network size of the current reconstruction algorithm.
We also plot the time its takes to reconstruct the network resistances (Fig.~\ref{fig:simu_result}a).
The time is on the order of milliseconds for network length smaller than $6$, and growth roughly to the square of the network length, hence proportional to the number of resistors.
This shows that our algorithm greatly outperforms previous results based on brute-force optimization~\cite{medina_2017_resistor-based}, which converges on the order of minutes.

The exponentially growing reconstruction error raises a concern on how practical is this algorithm for larger network size given noisy measurement.
We further simulate noisy measurement by multiplying each element in the noise-free $\Lambda$ a random number from a Gaussian distribution with mean equals $1$ and standard deviation $\sigma$.
We also include the symmetrization step on the noisy $\Lambda$ as we did in the experiments (Sec.\ref{sec:exp-setup}).
The RMSE as a function of noise level $\sigma$ for network length $4,\,7$ and $10$ are plotted in Fig.~\ref{fig:simu_result}(b).
The RMSEs grow linearly verse $\sigma$ for all network length tested, and if a RMSE of $0.1$ is desired, the corresponding maximum noise level is about $0.01$, $10^{-5}$, and $10^{-9}$, respectively.
Our experimental result in the unstretched state (Fig.~\ref{fig:exp_result}a) is consistent with the numerical simulation, as the ammeter measurement has an accuracy on the order of $10^{-3}$ and the repeated measurement for the unstretched state has errors about $10\%$.
Therefore, the presence of noise in $\Lambda$ measurement pose another upper limit on the size of accurately reconstructible resistor network based on our methods.

\section{DISCUSSION AND CONCLUSIONS}\label{sec:conclusion}
In this report, we develop a soft resistive strain sensor network and a measurement method to simultaneously acquire all of its resistor values from its peripheral nodes only.
\yz{The network has a total of $40$ independent resistive strain sensors.}
The observed relative change in resistances show patterns that are spatially correlated to the mechanical stretching applied to the sensor network, demonstrating its potential for detecting complex deformation patterns.
This method is based on the exact solution of the resistor network inverse problem~\cite{curtis_2000_inverse}, hence have a clear interpretation of its result and reconstruction process.
Through \yz{computer experiments}, we find the current reconstruction algorithm does not scale well with network size and is susceptible to measurement noise.
\yz{These two sources of error may limit the response speed, resolution or body coverage in practical soft robotic applications. Depending on the application demands, median speed applications or human-machine interface might be more suitable.}
%The current demonstrated network length is $4$ (or $5$ for the bench test) and further improvement on the reconstruction method is desired.

Developing soft sensors that can detect complex deformations in adaptive robotics is a challenging task.
Using a network of sensors distributed through the robot body could be a general solution to tackle this problem.
The ability to infer internal resistances from the boundary can reduce wiring needs in sensor fabrication, allow higher sensor surface density, and create thin and lightweight sensors for soft robots.
In the present work, our setup is limited by the simple sensor fabrication, the sampling rate of the ammeter and switching frequency of the eletro-magnetic relays, as well as the error of the reconstruction algorithm.
Nevertheless, the method is applicable to any resistor-type sensor network, such as bending, twisting, optical, temperature sensors, or even a hybrid of them for multimodal capacity.
In future works, we plan to improve the sensor fabrication method, increase the speed of processing, and apply this technology to pattern and object detections in robotic manipulator and wearable devices.
Also, different sensor network topology can be investigated~\cite{thuruthel_2020_joint}.

\balance

%%%%%%%%%%%%%%%%%%%%%%%%%%%%%%%%%%%%%%%%%%%%%%%%%%%%%%%%%%%%%%%%%%%%%%%%%%%%%%%%
%\section*{APPENDIX}

%\section*{ACKNOWLEDGMENT}
%We acknowledge the relevant funding xxx

%%%%%%%%%%%%%%%%%%%%%%%%%%%%%%%%%%%%%%%%%%%%%%%%%%%%%%%%%%%%%%%%%%%%%%%%%%%%%%%%

\bibliographystyle{IEEEtran}
\bibliography{IEEEabrv,mybibfile}

\begin{thebibliography}{10}
\providecommand{\url}[1]{#1}
\csname url@rmstyle\endcsname
\providecommand{\newblock}{\relax}
\providecommand{\bibinfo}[2]{#2}
\providecommand\BIBentrySTDinterwordspacing{\spaceskip=0pt\relax}
\providecommand\BIBentryALTinterwordstretchfactor{4}
\providecommand\BIBentryALTinterwordspacing{\spaceskip=\fontdimen2\font plus
\BIBentryALTinterwordstretchfactor\fontdimen3\font minus
  \fontdimen4\font\relax}
\providecommand\BIBforeignlanguage[2]{{%
\expandafter\ifx\csname l@#1\endcsname\relax
\typeout{** WARNING: IEEEtran.bst: No hyphenation pattern has been}%
\typeout{** loaded for the language `#1'. Using the pattern for}%
\typeout{** the default language instead.}%
\else
\language=\csname l@#1\endcsname
\fi
#2}}

\bibitem{rus_2015_design}
\BIBentryALTinterwordspacing
D.~Rus and M.~T. Tolley, ``Design, fabrication and control of soft robots,''
  \emph{Nature}, vol. 521, no. 7553, pp. 467--475, May 2015, number: 7553.
  [Online]. Available: \url{https://doi.org/10.1038/nature14543}
\BIBentrySTDinterwordspacing

\bibitem{shintake_2018_soft}
J.~Shintake, V.~Cacucciolo, D.~Floreano, and H.~Shea,
  ``\BIBforeignlanguage{en}{Soft {Robotic} {Grippers}},''
  \emph{\BIBforeignlanguage{en}{Advanced Materials}}, vol.~30, no.~29, p.
  1707035, 2018.

\bibitem{goh_2022_3d}
G.~D. Goh, G.~L. Goh, Z.~Lyu, M.~Z. Ariffin, W.~Y. Yeong, G.~Z. Lum,
  D.~Campolo, B.~S. Han, and H.~Y.~A. Wong, ``\BIBforeignlanguage{en}{{3D}
  {Printing} of {Robotic} {Soft} {Grippers}: {Toward} {Smart} {Actuation} and
  {Sensing}},'' \emph{\BIBforeignlanguage{en}{Advanced Materials
  Technologies}}, vol. n/a, no. n/a, p. 2101672, 2022.

\bibitem{shepherd_2011_multigait}
\BIBentryALTinterwordspacing
R.~F. Shepherd, F.~Ilievski, W.~Choi, S.~A. Morin, A.~A. Stokes, A.~D. Mazzeo,
  X.~Chen, M.~Wang, and G.~M. Whitesides, ``\BIBforeignlanguage{en}{Multigait
  soft robot},'' \emph{\BIBforeignlanguage{en}{Proceedings of the National
  Academy of Sciences}}, vol. 108, no.~51, pp. 20\,400--20\,403, Dec. 2011,
  publisher: Proceedings of the National Academy of Sciences. [Online].
  Available: \url{https://www.pnas.org/doi/abs/10.1073/pnas.1116564108}
\BIBentrySTDinterwordspacing

\bibitem{brown_2010_universal}
E.~Brown, N.~Rodenberg, J.~Amend, A.~Mozeika, E.~Steltz, M.~R. Zakin,
  H.~Lipson, and H.~M. Jaeger, ``Universal robotic gripper based on the jamming
  of granular material,'' \emph{Proceedings of the National Academy of
  Sciences}, vol. 107, no.~44, pp. 18\,809--18\,814, Nov. 2010.

\bibitem{tolley_2014_resilient}
\BIBentryALTinterwordspacing
M.~T. Tolley, R.~F. Shepherd, B.~Mosadegh, K.~C. Galloway, M.~Wehner,
  M.~Karpelson, R.~J. Wood, and G.~M. Whitesides, ``A {Resilient}, {Untethered}
  {Soft} {Robot},'' \emph{Soft Robotics}, vol.~1, no.~3, pp. 213--223, 2014,
  publisher: Mary Ann Liebert, Inc., publishers. [Online]. Available:
  \url{https://www.liebertpub.com/doi/10.1089/soro.2014.0008}
\BIBentrySTDinterwordspacing

\bibitem{meerbeek_2018_soft}
\BIBentryALTinterwordspacing
I.~M.~V. Meerbeek, C.~M.~D. Sa, and R.~F. Shepherd,
  ``\BIBforeignlanguage{EN}{Soft optoelectronic sensory foams with
  proprioception},'' \emph{\BIBforeignlanguage{EN}{Science Robotics}}, Nov.
  2018, publisher: American Association for the Advancement of Science.
  [Online]. Available:
  \url{https://www.science.org/doi/abs/10.1126/scirobotics.aau2489}
\BIBentrySTDinterwordspacing

\bibitem{hughes_2018_tactile}
\BIBentryALTinterwordspacing
J.~Hughes and F.~Iida, ``Tactile {Sensing} {Applied} to the {Universal}
  {Gripper} {Using} {Conductive} {Thermoplastic} {Elastomer},'' \emph{Soft
  Robotics}, vol.~5, no.~5, pp. 512--526, 2018, publisher: Mary Ann Liebert,
  Inc., publishers. [Online]. Available:
  \url{https://www.liebertpub.com/doi/abs/10.1089/soro.2017.0089}
\BIBentrySTDinterwordspacing

\bibitem{zhang_2021_fruit}
\BIBentryALTinterwordspacing
J.~Zhang, S.~Lai, H.~Yu, E.~Wang, X.~Wang, and Z.~Zhu,
  ``\BIBforeignlanguage{en}{Fruit {Classification} {Utilizing} a {Robotic}
  {Gripper} with {Integrated} {Sensors} and {Adaptive} {Grasping}},''
  \emph{\BIBforeignlanguage{en}{Mathematical Problems in Engineering}}, vol.
  2021, p. e7157763, Sept. 2021, publisher: Hindawi. [Online]. Available:
  \url{https://www.hindawi.com/journals/mpe/2021/7157763/}
\BIBentrySTDinterwordspacing

\bibitem{souri_2020_wearable}
H.~Souri, H.~Banerjee, A.~Jusufi, N.~Radacsi, A.~A. Stokes, I.~Park, M.~Sitti,
  and M.~Amjadi, ``\BIBforeignlanguage{en}{Wearable and {Stretchable} {Strain}
  {Sensors}: {Materials}, {Sensing} {Mechanisms}, and {Applications}},''
  \emph{\BIBforeignlanguage{en}{Advanced Intelligent Systems}}, vol.~2, no.~8,
  p. 2000039, 2020.

\bibitem{shih_2020_electronic}
\BIBentryALTinterwordspacing
B.~Shih, D.~Shah, J.~Li, T.~G. Thuruthel, Y.-L. Park, F.~Iida, Z.~Bao,
  R.~Kramer-Bottiglio, and M.~T. Tolley, ``\BIBforeignlanguage{EN}{Electronic
  skins and machine learning for intelligent soft robots},''
  \emph{\BIBforeignlanguage{EN}{Science Robotics}}, Apr. 2020, publisher:
  American Association for the Advancement of Science. [Online]. Available:
  \url{https://www.science.org/doi/10.1126/scirobotics.aaz9239}
\BIBentrySTDinterwordspacing

\bibitem{castano_2014_smart}
\BIBentryALTinterwordspacing
L.~M. Castano and A.~B. Flatau, ``\BIBforeignlanguage{en}{Smart fabric sensors
  and e-textile technologies: a review},'' \emph{\BIBforeignlanguage{en}{Smart
  Materials and Structures}}, vol.~23, no.~5, p. 053001, 2014, publisher: IOP
  Publishing. [Online]. Available:
  \url{https://doi.org/10.1088/0964-1726/23/5/053001}
\BIBentrySTDinterwordspacing

\bibitem{buckner_2020_roboticizing}
T.~L. Buckner, R.~A. Bilodeau, S.~Y. Kim, and R.~Kramer-Bottiglio,
  ``Roboticizing fabric by integrating functional fibers,'' \emph{Proceedings
  of the National Academy of Sciences}, vol. 117, no.~41, pp. 25\,360--25\,369,
  Oct. 2020.

\bibitem{nittala_2018_multi-touch}
A.~S. Nittala, A.~Withana, N.~Pourjafarian, and J.~Steimle, ``Multi-{Touch}
  {Skin}: {A} {Thin} and {Flexible} {Multi}-{Touch} {Sensor} for {On}-{Skin}
  {Input},'' in \emph{Proceedings of the 2018 {CHI} {Conference} on {Human}
  {Factors} in {Computing} {Systems}}, ser. {CHI} '18.\hskip 1em plus 0.5em
  minus 0.4em\relax New York, NY, USA: Association for Computing Machinery,
  2018, pp. 1--12.

\bibitem{hua_2018_skin-inspired}
\BIBentryALTinterwordspacing
Q.~Hua, J.~Sun, H.~Liu, R.~Bao, R.~Yu, J.~Zhai, C.~Pan, and Z.~L. Wang,
  ``\BIBforeignlanguage{en}{Skin-inspired highly stretchable and conformable
  matrix networks for multifunctional sensing},''
  \emph{\BIBforeignlanguage{en}{Nature Communications}}, vol.~9, no.~1, p. 244,
  2018, number: 1 Publisher: Nature Publishing Group. [Online]. Available:
  \url{https://www.nature.com/articles/s41467-017-02685-9}
\BIBentrySTDinterwordspacing

\bibitem{pinto_2017_cnt-based}
\BIBentryALTinterwordspacing
T.~Pinto, L.~Cai, C.~Wang, and X.~Tan, ``\BIBforeignlanguage{en}{{CNT}-based
  sensor arrays for local strain measurements in soft pneumatic actuators},''
  \emph{\BIBforeignlanguage{en}{International Journal of Intelligent Robotics
  and Applications}}, vol.~1, no.~2, pp. 157--166, 2017. [Online]. Available:
  \url{https://doi.org/10.1007/s41315-017-0018-6}
\BIBentrySTDinterwordspacing

\bibitem{koivikko_2018_screen-printed}
A.~Koivikko, E.~Sadeghian~Raei, M.~Mosallaei, M.~Mäntysalo, and V.~Sariola,
  ``Screen-{Printed} {Curvature} {Sensors} for {Soft} {Robots},'' \emph{IEEE
  Sensors Journal}, vol.~18, no.~1, pp. 223--230, 2018, conference Name: IEEE
  Sensors Journal.

\bibitem{lee_2016_transparent}
\BIBentryALTinterwordspacing
S.~Lee, A.~Reuveny, J.~Reeder, S.~Lee, H.~Jin, Q.~Liu, T.~Yokota, T.~Sekitani,
  T.~Isoyama, Y.~Abe, Z.~Suo, and T.~Someya, ``\BIBforeignlanguage{en}{A
  transparent bending-insensitive pressure sensor},''
  \emph{\BIBforeignlanguage{en}{Nature Nanotechnology}}, vol.~11, no.~5, pp.
  472--478, May 2016, number: 5 Publisher: Nature Publishing Group. [Online].
  Available: \url{http://www.nature.com/articles/nnano.2015.324}
\BIBentrySTDinterwordspacing

\bibitem{georgopoulou_2022_pellet-based}
\BIBentryALTinterwordspacing
A.~Georgopoulou and F.~Clemens, ``\BIBforeignlanguage{en}{Pellet-based fused
  deposition modeling for the development of soft compliant robotic grippers
  with integrated sensing elements},'' \emph{\BIBforeignlanguage{en}{Flexible
  and Printed Electronics}}, vol.~7, no.~2, p. 025010, 2022, publisher: IOP
  Publishing. [Online]. Available:
  \url{https://doi.org/10.1088/2058-8585/ac6f34}
\BIBentrySTDinterwordspacing

\bibitem{pourjafarian_2019_multi-touch}
\BIBentryALTinterwordspacing
N.~Pourjafarian, A.~Withana, J.~A. Paradiso, and J.~Steimle, ``Multi-{Touch}
  {Kit}: {A} {Do}-{It}-{Yourself} {Technique} for {Capacitive} {Multi}-{Touch}
  {Sensing} {Using} a {Commodity} {Microcontroller},'' in \emph{Proceedings of
  the 32nd {Annual} {ACM} {Symposium} on {User} {Interface} {Software} and
  {Technology}}, ser. {UIST} '19.\hskip 1em plus 0.5em minus 0.4em\relax New
  York, NY, USA: Association for Computing Machinery, 2019, pp. 1071--1083,
  event-place: New Orleans, LA, USA. [Online]. Available:
  \url{https://doi.org.remotexs.ntu.edu.sg/10.1145/3332165.3347895}
\BIBentrySTDinterwordspacing

\bibitem{audenaert_2015_granual}
L.~E.~G. Audenaert, L.~Van~Bremen, and D.~S. Hulsinga,
  ``\BIBforeignlanguage{en}{Granual {Jamming} {Gripper}: {The} {Design} of a
  {Universal} {Gripper} {And} {Accompanying} {Capacitive} {Sensor} {Array}},''
  2015.

\bibitem{loh_2021_3d}
L.~Y.~W. Loh, U.~Gupta, Y.~Wang, C.~C. Foo, J.~Zhu, and W.~F. Lu,
  ``\BIBforeignlanguage{en}{{3D} {Printed} {Metamaterial} {Capacitive}
  {Sensing} {Array} for {Universal} {Jamming} {Gripper} and {Human} {Joint}
  {Wearables}},'' \emph{\BIBforeignlanguage{en}{Advanced Engineering
  Materials}}, vol.~23, no.~5, p. 2001082, 2021.

\bibitem{hughes_2021_online}
\BIBentryALTinterwordspacing
J.~Hughes, L.~Scimeca, P.~Maiolino, and F.~Iida, ``Online {Morphological}
  {Adaptation} for {Tactile} {Sensing} {Augmentation},'' \emph{Frontiers in
  Robotics and AI}, vol.~8, p. 665030, July 2021. [Online]. Available:
  \url{https://www.ncbi.nlm.nih.gov/pmc/articles/PMC8329453/}
\BIBentrySTDinterwordspacing

\bibitem{medina_2017_resistor-based}
O.~Medina, A.~Shapiro, and N.~Shvalb, ``Resistor-{Based} {Shape} {Sensor} for a
  {Spatial} {Flexible} {Manifold},'' \emph{IEEE Sensors Journal}, vol.~17,
  no.~1, pp. 46--50, 2017, conference Name: IEEE Sensors Journal.

\bibitem{curtis_2000_inverse}
\BIBentryALTinterwordspacing
E.~B. Curtis and J.~A. Morrow, \emph{Inverse {Problems} for {Electrical}
  {Networks}}, ser. Series on {Applied} {Mathematics}.\hskip 1em plus 0.5em
  minus 0.4em\relax WORLD SCIENTIFIC, Mar. 2000, vol. Volume 13, no. Volume 13.
  [Online]. Available: \url{https://doi.org/10.1142/4306}
\BIBentrySTDinterwordspacing

\bibitem{oberlin_2000_discrete}
R.~Oberlin, ``Discrete inverse problems for {Schrödinger} and resistor
  networks,'' \emph{Research archive of Research Experiences for Undergraduates
  program at Univ. of Washington}, 2000.

\bibitem{borcea_2008_electrical}
\BIBentryALTinterwordspacing
L.~Borcea, V.~Druskin, and F.~G. Vasquez, ``\BIBforeignlanguage{en}{Electrical
  impedance tomography with resistor networks},''
  \emph{\BIBforeignlanguage{en}{Inverse Problems}}, vol.~24, no.~3, p. 035013,
  2008, publisher: IOP Publishing. [Online]. Available:
  \url{https://doi.org/10.1088/0266-5611/24/3/035013}
\BIBentrySTDinterwordspacing

\bibitem{park2022biomimetic}
K.~Park, H.~Yuk, M.~Yang, J.~Cho, H.~Lee, and J.~Kim, ``A biomimetic
  elastomeric robot skin using electrical impedance and acoustic tomography for
  tactile sensing,'' \emph{Science Robotics}, vol.~7, no.~67, p. eabm7187,
  2022.

\bibitem{shih2020electronic}
B.~Shih, D.~Shah, J.~Li, T.~G. Thuruthel, Y.-L. Park, F.~Iida, Z.~Bao,
  R.~Kramer-Bottiglio, and M.~T. Tolley, ``Electronic skins and machine
  learning for intelligent soft robots,'' \emph{Science Robotics}, vol.~5,
  no.~41, p. eaaz9239, 2020.

\bibitem{thuruthel_2020_joint}
T.~G. Thuruthel, J.~Hughes, and F.~Iida, ``Joint {Entropy}-{Based} {Morphology}
  {Optimization} of {Soft} {Strain} {Sensor} {Networks} for {Functional}
  {Robustness},'' \emph{IEEE Sensors Journal}, vol.~20, no.~18, pp.
  10\,801--10\,810, Sept. 2020, conference Name: IEEE Sensors Journal.

\end{thebibliography}


%%%%%%%%%%%%%%%%%%%%%%%%%%%%%%%%%%%%%%%%%%%%%%%%%%%%%%%%%%%%%%%%%%%%%%%%%%%%%%%%

%%%%%%%%%%%%%%%%%%%%%%%%%%%%%%%%%%%%%%%%%%%%%%%%%%%%%%%%%%%%%%%%%%%%%%%%%%%%%%%%

\end{document}